\documentclass{article}

\usepackage[final]{tackling_climate_workshop_style}
\usepackage[utf8]{inputenc} 
\usepackage[T1]{fontenc}    
\usepackage[hypertexnames=false]{hyperref} 
\usepackage{url}            
\usepackage{booktabs}       
\usepackage{amsfonts}       
\usepackage{nicefrac}       
\usepackage{microtype}      
\usepackage{amsmath}
\usepackage{xcolor}
\usepackage{float}
\usepackage{placeins}
\usepackage{multirow}
\usepackage{enumitem}  
\usepackage{calc}
\usepackage{graphicx}
\usepackage{appendix}
\usepackage{mhchem}

\title{Data-driven Optimal Sensor Placement for Aerosol Monitoring in the UK}

\author{
\begin{minipage}[t]{\textwidth}
\centering
\begin{tabular}{ccc}
\textbf{Talha Ansar} & \textbf{Waqar Muhammad Ashraf}$^*$ & \textbf{Alejandro Coca-Castro} \\
University of Engineering & The Alan Turing Institute  & The Alan Turing Institute\\
and Technology & London, UK & London, UK \\
Lahore, Pakistan  \\
\texttt{} & \texttt{washraf@turing.ac.uk} & \texttt{} \\
\end{tabular}
\end{minipage}
\vspace{1.5em}
\\
\begin{minipage}[t]{\textwidth}
\centering
\begin{tabular}{ccc}
\textbf{Manvendra Janmaijaya } & \textbf{Ramit Debnath} & \textbf{J. Scott Hosking} \\
\textbf{The Alan Turing Institute} & \textbf{University of Cambridge} & \textbf{The Alan Turing Institute} \\
\textbf{London, UK} & \textbf{Cambridge, UK} & \textbf{London, UK} \\
\texttt{} & \texttt{} & \texttt{} \\
\end{tabular}
\end{minipage}
}

\begin{document}
\maketitle

\begin{abstract}
We develop a data-driven sensor-placement framework and apply it to aerosol monitoring in the United Kingdom (UK). A trained convolutional Gaussian neural process is embedded within the DeepSensor optimisation framework. Genetic optimisation with lazy evaluation and caching reduces mean computational time by 60\% (10$\pm$3 versus 25$\pm$7 minutes) while producing sequential placement performance comparable to the greedy solver. Using the CAMS reference field to construct a retrospective oracle benchmark, Joint NLL improves by about 22\% for Joint-MI-driven genetic placement after five locations for aerosol stations are identified. Seasonal and wind-regime analyses show that agreement between model-based criteria and oracle benchmarks varies with meteorological transport conditions. These results show a computationally efficient approach for data-informed expansion of environmental observing systems.
\end{abstract}

\section{Introduction}

Atmospheric aerosols influence the radiative balance of Earth and contribute to adverse human-health outcomes \cite{ipcc2021wg1,cohen2017gbd}. Expanding observation networks for aerosol monitoring is therefore increasingly important, but sensor placement over large spatial domains can be computationally costly under practical deployment constraints \cite{vandonkelaar2015pm25}. Atmospheric transport phenomena add a further challenge since changing circulation dynamics redistributes aerosol concentrations and reshapes their spatial gradients. Probabilistic machine-learning models can provide efficient predictions together with uncertainty estimates \cite{rasmussen2006gp,adams2007bocpd}. Convolutional Gaussian neural processes (ConvGNPs) encode observations on spatial grids and enable linear-scaling inference over large domains \cite{gordon2020cnp}. We therefore examine the effectiveness of a ConvGNP for aerosol sensor placement over the UK and quantify how meteorological variability affects agreement between model-based placement criteria and realised predictive benefit \cite{lorenc2014forecast,cucurull2024osse}.

This paper makes four contributions to data-driven aerosol sensor placement: (i) we embed a pre-trained ConvGNP aerosol model within the DeepSensor optimisation framework and estimate five new sensor locations across the UK \cite{deepsensor,pelucchi2023poster}, (ii) compare genetic optimisation with lazy evaluation and caching against exhaustive greedy search, (iii) compare the performance trajectories and final geographic placement profiles produced by model-based criteria and retrospective oracle benchmarks, and (iv) investigate how seasonal and wind-regime variation affects agreement between model-based placement criteria and realised predictive benefit.

\subsection{Prior work}
Probabilistic environmental sensor placement has traditionally used Gaussian processes to combine sparse observations with predictive uncertainty \cite{rasmussen2006gp}. Neural-process models extend this approach to large spatial grids while retaining uncertainty quantification \cite{gordon2020cnp}. Andersson et al.\ \cite{andersson2023convgnp} applied a ConvGNP to Antarctic thermal-anomaly monitoring but did not examine seasonal variation in sensor placement accuracy. Pelucchi \cite{pelucchi2023poster} developed a ConvGNP workflow for black carbon aerosol optical depth across Europe and evaluated several acquisition functions using greedy search. Chapron et al.\ \cite{chapron2026optimal} used Gumbel-Softmax optimisation for ocean-state monitoring, with simultaneous placement and Monte Carlo evaluation. Against this broader optimisation literature, the present study focuses on two narrower questions: how genetic search compares with exhaustive greedy evaluation within a ConvGNP-based aerosol placement workflow, and how agreement between model-based placement criteria and retrospective oracle benchmarks varies across seasonal cycles and wind regimes.

\section{Methods}
The workflow combines prospective model-based placement with retrospective evaluation. The ConvGNP first generates probabilistic predictions of black carbon aerosol optical depth from the current observation context. Candidate locations are scored by Joint Mutual Information (Joint MI), Marginal MI, and Mean Variance (Mean Var) from the model predictive distribution, favouring locations that are expected to be informative or reduce predictive uncertainty. The Copernicus Atmosphere Monitoring Service (CAMS) reference field is available for the historical study period, the value at each candidate can also be revealed retrospectively, appended to the context, and the resulting change in predictive performance measured using metrics such as Joint Negative Log-Likelihood (Joint NLL), Marginal Negative Log-Likelihood (Marginal NLL), and Root Mean Square Error (RMSE). These reference-based scores form oracle benchmarks that are unavailable during a real deployment but provide a direct measure of the realised benefit of observing each candidate location.

\subsection{Data, model, and experimental settings}
A pretrained ConvGNP implemented in DeepSensor generates probabilistic predictions of black carbon aerosol optical depth on a spatiotemporal grid \cite{deepsensor,pelucchi2023poster}. The model was trained at European scale using CAMS reanalysis and is applied here to the UK subset of the original domain. AERONET stations define the initial observation context \cite{holben1998aeronet}. CAMS provides gridded black carbon aerosol optical depth and near-surface winds ($u_{10}$ and $v_{10}$) \cite{inness2019cams}. Candidate locations are CAMS grid cells restricted to land using a land-sea mask $\mathrm{(lsm)}$ threshold $>$ 0.5. We evaluate the accuracy of sequential placement of five sensors from the test dataset of 2017 to 2019 duration, sampled at 10-day intervals.

\subsection{Optimisation setup}
Evaluating sensor-placement acquisition functions over large spatiotemporal grids is computationally expensive, while joint multi-station placement does not closely reflect practical deployment. Sensors are commonly installed incrementally because of procurement, permitting, site access, and maintenance constraints. We therefore select one location at a time and update the network after each installation. This repeated single-step strategy approximates multi-station planning and follows active-learning sensor-placement practice \cite{andersson2023convgnp,pelucchi2023poster}.

Let $\mathcal{C}_k$ denote the context set after $k$ new sensors have been added, consisting of installed sensors plus the first $k$ proposed sensors. Let $\mathcal{X}$ denote the set of feasible candidate locations. In each placement step $k\in\{0,\dots,N_{\text{new}}-1\}$, the next location is selected as
\begin{equation}
 x_{k+1} \;=\; \arg\max_{x \in \mathcal{X}\setminus\{x_1,\dots,x_k\}} \; a(x;\mathcal{C}_k),
\label{eq:greedy_generic}
\end{equation}
where $a(x;\mathcal{C}_k)$ denotes the acquisition score assigned to candidate location $x$ given the current context $\mathcal{C}_k$. For prospective placement, the score is obtained from the ConvGNP predictive distribution using Joint MI, Marginal MI, and Mean Var. After selecting $x_{k+1}$, an observation at that location is appended to the context set and the procedure is repeated. Revealing the CAMS value at a candidate location allows the realised change in Joint NLL, Marginal NLL, or RMSE to be computed, producing the corresponding oracle score. Formal definitions of the model-based and oracle criteria are provided in Appendix~\ref{app:acquisition_functions}.

\subsubsection{Greedy and Genetic Optimisation solvers}
We apply genetic and greedy solvers for optimising the sensor placement site with Eq.~\ref{eq:greedy_generic}. The greedy solver evaluates every feasible candidate sensor location at each placement step and selects the best-scoring location \cite{andersson2023convgnp,pelucchi2023poster}. The genetic solver represents a proposed location by two integer genes, $[i_{\mathrm{lat}},i_{\mathrm{lon}}]$, corresponding to latitude and longitude grid indices. The genetic search spans the complete rectangular latitude--longitude grid, while masked sea cells and locations selected in earlier placement steps receive a large fitness penalty. Uniform crossover can therefore recombine the latitude and longitude components of different parent locations, while mutation changes the coordinate genes within their permitted index ranges. Candidate evaluations are cached within each placement step. When an acquisition function supports parallel evaluation, the complete feasible score surface is computed once and reused as a lookup table. Otherwise, candidate scores are evaluated lazily, meaning that an expensive acquisition score is computed only when the genetic search visits a feasible coordinate pair for the first time, and subsequent visits reuse the cached value. Further implementation details are provided in Appendix~\ref{app:ga_details}.

\subsection{Weather-regime agreement analysis}
To quantify whether model-based criteria identify the same high-value regions as their oracle counterparts, we compute Spearman rank correlations between paired candidate-score surfaces: Joint MI with Joint NLL, Marginal MI with Marginal NLL, and Mean Var with RMSE. The oracle metrics are sign-aligned so that larger scores correspond to greater realised benefit, making positive correlation equivalent to agreement in candidate ranking. Spearman correlation captures rank agreement and is insensitive to monotonic rescaling. Evaluation dates are stratified into north, east, south, and west sectors using domain-mean near-surface wind direction derived from CAMS $u_{10}$ and $v_{10}$. Monthly sector frequencies are then used to relate acquisition-function agreement to seasonal transport regimes.

\section{Results}
\subsection{Sequential placement performance and computational efficiency}
\begin{figure}[h]
  \centering
  \includegraphics[width=0.65\linewidth]{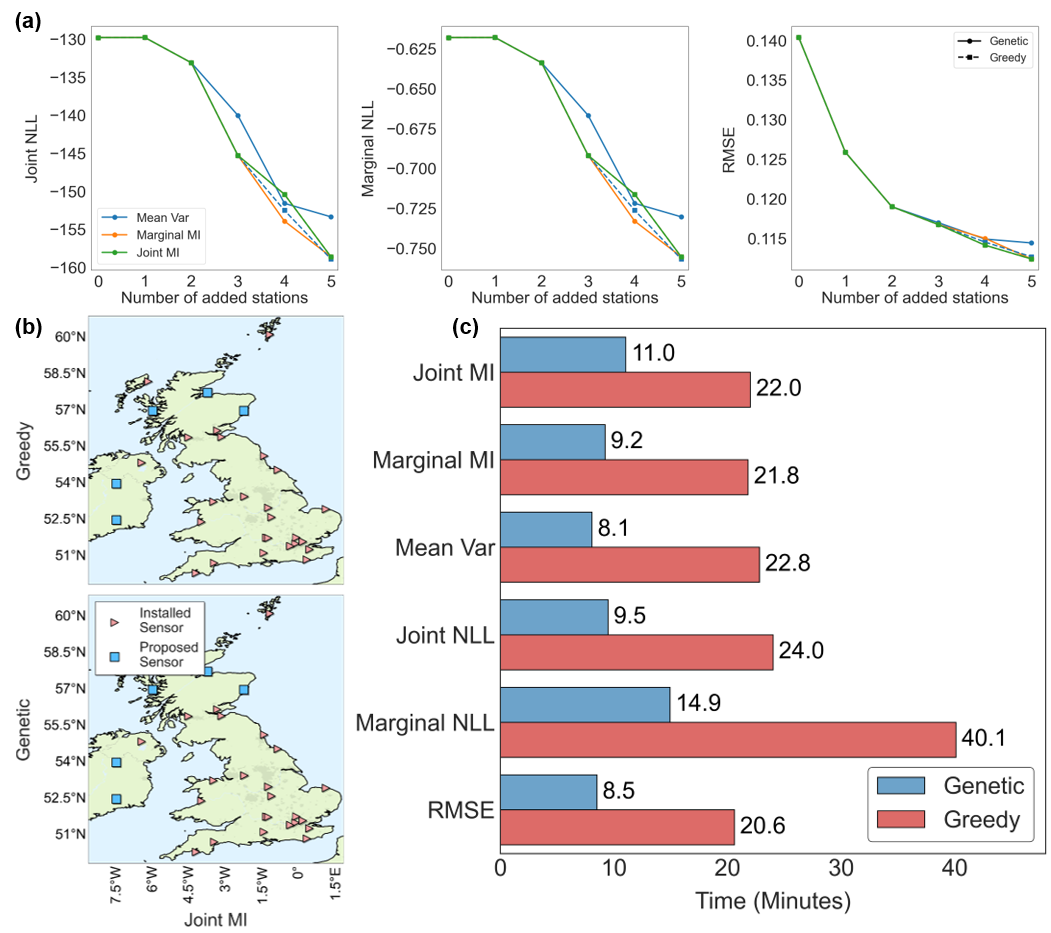}
  \caption{Sequential sensor placement performance with optimisation solvers. (a) Joint NLL, Marginal NLL, and RMSE as five stations are added using the model-based acquisition criteria. (b) Joint-MI-selected locations for greedy and genetic search. (c) Wall-clock time required to identify five sensor locations for each acquisition criterion.}
  \label{fig:main_summary}
\end{figure}

Figure~\ref{fig:main_summary} summarises sequential placement performance, representative placement geography, and runtime. Figure~\ref{fig:main_summary}(a) evaluates the model-based aerosol sensor placement sequences using Joint NLL, Marginal NLL, and RMSE. All three lower-is-better metrics improve as five sensors are added in the existing network of sensors. For Joint-MI-driven genetic placement, Joint NLL decreases from about $-130$ to about $-159$, corresponding to an improvement of approximately 22\%. Across the evaluation criteria and solvers, Marginal NLL decreases from about $-0.62$ to between approximately $-0.76$ and $-0.73$, while RMSE decreases from about $0.140$ to approximately $0.113$--$0.115$. The closely matched trajectories show that the genetic and greedy solvers retain similar sequential predictive performance despite selecting different grid cells.

Figure~\ref{fig:main_summary}(b) shows the five Joint-MI-selected sensor locations for greedy and genetic search. Both solvers retain broad UK coverage while differing at several grid cells. Full placement maps for the three non-oracle acquisition criteria and their oracle variants are provided in Appendix~\ref{fig:app_placements}. Figure~\ref{fig:main_summary}(c) shows that genetic search is faster than greedy search for all six acquisition criteria. Mean runtime decreases from 25$\pm$7 minutes to 10$\pm$3 minutes, corresponding to a 60\% reduction. The largest absolute saving occurs for Marginal NLL, where runtime decreases from 40.1 to 14.9 minutes. The computational saving accumulates because each candidate score is aggregated over the multi-year evaluation tasks and the search is repeated across five sequential placement steps and multiple acquisition criteria. The reduced runtime is obtained while retaining the closely matched predictive-performance trajectories shown in Figure~\ref{fig:main_summary}(a).

\subsection{Weather dependence of acquisition-function agreement}
\begin{figure}[h]
  \centering
  \includegraphics[width=0.6\linewidth]{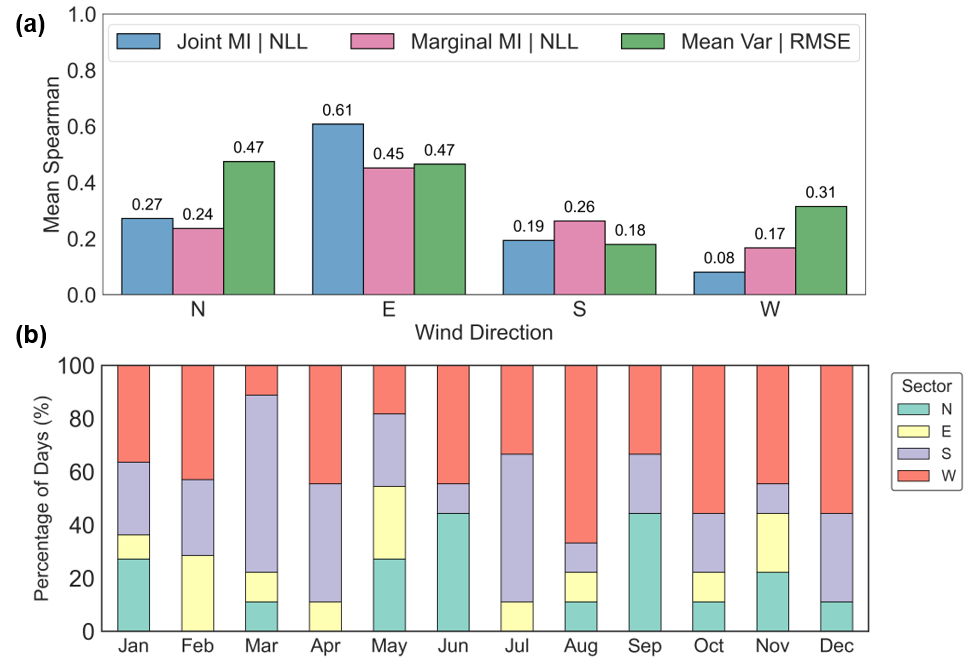}
  \caption{Wind-regime dependence of acquisition-function agreement. (a) Mean Spearman correlation between paired model-based and oracle acquisition functions, stratified by domain-mean wind direction. The pairs are Joint MI with Joint NLL, Marginal MI with Marginal NLL, and Mean Var with RMSE. (b) Monthly percentage of evaluation days assigned to north (N), east (E), south (S), and west (W) wind sectors.}
  \label{fig:weather_main}
\end{figure}

Figure~\ref{fig:weather_main}(a) shows that agreement between model-based criteria and their oracle counterparts varies with wind direction. The monthly distribution of wind sectors in Figure~\ref{fig:weather_main}(b) provides the seasonal context, with westerly and southerly regimes dominating several months and easterly episodes more prominent in February, May, and November. Under northerly winds, mean Spearman correlations are 0.27 for Joint MI with Joint NLL, 0.24 for Marginal MI with Marginal NLL, and 0.47 for Mean Var with RMSE. Easterly winds give correlations of 0.61, 0.45, and 0.47 for the same three pairs, respectively. Agreement is weaker under southerly winds (0.19, 0.26, and 0.18) and, for the information-based pairs, under westerly winds (0.08 and 0.17). These correlations are generally modest rather than indicating one-to-one correspondence between model uncertainty and realised predictive benefit. Their systematic variation across wind regimes shows that the usefulness of uncertainty-based placement criteria changes with atmospheric transport conditions.

\section{Conclusion}
Genetic and greedy solvers are used for sequential placement of five new sensor locations for aerosol monitoring across the UK. Genetic search reduces mean runtime from 25$\pm$7 to 10$\pm$3s minutes, a 60\% reduction, while maintaining closely matched sequential predictive performance with the greedy approach. For Joint-MI-driven genetic placement, Joint NLL improves by approximately 22\% after five sensors are added to the existing network. Acquisition-landscape agreement also varies with meteorological conditions: under easterly winds, mean Spearman correlations reach 0.61 for Joint MI with Joint NLL and 0.45 for Marginal MI with Marginal NLL.  The modest correlations show that model-based uncertainty is not a direct substitute for realised predictive improvement, while the regime dependence indicates that atmospheric transport affects when the model-based criteria most closely identify high-value observation locations. The resulting framework combines probabilistic prediction, comparative computationally efficient genetic search, and retrospective performance benchmarking for data-driven expansion of environmental observing networks. In conclusion, ConvGNPs can guide sequential sensor placement, while retrospective benchmarks verify whether uncertainty criteria match realised predictive gains. As rank correlation varies with wind direction, candidate sensor sites should be assessed across distinct atmospheric transport regimes rather than aggregate scores alone. Evaluating the developed framework on other pollutants and environmental variables across the regions is left to future work.

\section*{Acknowledgments}

Waqar Muhammad Ashraf gratefully acknowledges that this work was conducted during his appointment as a Visiting Scientist within the Environmental Forecasting Mission at The Alan Turing Institute, UK. Talha Ansar would like to thank Anupam Srivastava for invaluable mentorship provided through the Climate Change AI Mentorship Program. In addition, the authors express their gratitude to Jack Richings (Research Data Scientist - The Alan Turing Institute) for technical feedback and valuable insights regarding the implementation of the genetic algorithm. The authors acknowledge the use of Anthropic's Claude to refine grammar, improve expression, and enhance the clarity of the arguments.
\bibliographystyle{unsrt}
\bibliography{reference}
\clearpage

\begin{appendices}
\renewcommand{\thefigure}{A.\arabic{figure}}
\renewcommand{\thetable}{A.\arabic{table}}
\renewcommand{\theHfigure}{A.\arabic{figure}}
\renewcommand{\theHtable}{A.\arabic{table}}
\setcounter{figure}{0}
\setcounter{table}{0}
\appendix
\section{Appendix}

\subsection{Acquisition functions}
\label{app:acquisition_functions}
Acquisition functions are evaluated over the candidate grid and aggregated across tasks. The three model-based criteria are Joint Mutual Information (Joint MI), Marginal Mutual Information (Marginal MI), and Mean Variance (Mean Var). Joint and Marginal MI favour candidate observations that reduce uncertainty in the joint or marginal predictive distribution, while Mean Var reflects spatially averaged predictive variance. The three reference-based oracle criteria are Joint Negative Log-Likelihood (Joint NLL), Marginal Negative Log-Likelihood (Marginal NLL), and Root Mean Squared Error (RMSE). The oracle construction follows Andersson et al.~\cite{andersson2023convgnp}: for each candidate, the CAMS reference value is revealed, appended to the current context, and predictive performance over the target field is re-evaluated. The resulting change in NLL or RMSE represents the realised predictive benefit that would have followed from observing that location.

For a task at time $t$, let $p_\theta(\mathbf{y}_t \mid \mathbf{X}_t, \mathcal{C}_k)$ denote the model predictive distribution over a target set $\mathbf{X}_t$. For a prospective candidate $x$, let $\hat y_t(x)$ be the model-predicted mean at that location and $\mathcal{C}_k^x=\mathcal{C}_k\cup\{(x,\hat y_t(x))\}$ the candidate-augmented context. DeepSensor minimises joint entropy after this mean-value infill; the equivalent maximisation score used for Joint MI is
\begin{equation}
 a_{\mathrm{JMI},t}(x;\mathcal{C}_k) \;=\;
 H\!\left[p_\theta(\mathbf{y}_t\mid\mathbf{X}_t,\mathcal{C}_k)\right]
 -H\!\left[p_\theta(\mathbf{y}_t\mid\mathbf{X}_t,\mathcal{C}_k^x)\right],
\label{eq:joint_mi}
\end{equation}
where $H[\cdot]$ is differential entropy. In the ConvGNP, replacing the unknown candidate value with $\hat y_t(x)$ approximates the expectation over possible observations in mutual information.

Assuming a Gaussian marginal at each target point, the marginal entropy is:
\begin{equation}
 H_i(x,t) \;=\; \frac{1}{2}\log\!\left(2\pi e\,\sigma_{t,i}^2(x)\right),
\end{equation}
where $\sigma_{t,i}^2(x)$ is the variance at target $i$ under $\mathcal{C}_k^x$. Let $H_i^0(t)$ denote the corresponding entropy under $\mathcal{C}_k$. Marginal MI uses the mean reduction in marginal entropy:
\begin{equation}
 a_{\mathrm{MMI},t}(x;\mathcal{C}_k) \;=\; \frac{1}{|\mathbf{X}_t|}\sum_{i=1}^{|\mathbf{X}_t|}\left[H_i^0(t)-H_i(x,t)\right].
\label{eq:marginal_mi}
\end{equation}

Mean Var uses the reduction in spatially averaged predictive variance, where $\sigma_{t,i}^{2,0}$ is the variance under $\mathcal{C}_k$:
\begin{equation}
 a_{\mathrm{MV},t}(x;\mathcal{C}_k) \;=\; \frac{1}{|\mathbf{X}_t|}\sum_{i=1}^{|\mathbf{X}_t|}\left[\sigma_{t,i}^{2,0}-\sigma_{t,i}^2(x)\right].
\label{eq:mean_var}
\end{equation}
For a fixed task, the pre-augmentation terms in these three scores are constant across candidates. Maximising the reductions gives the same candidate rankings as DeepSensor's minimisation of the post-augmentation joint entropy, mean marginal entropy, and mean variance.

To make the oracle construction explicit, let $\mathbf{y}_t^{\star}$ denote CAMS reference values over the target set and $y_t^{\star}(x)$ the CAMS value at candidate $x$. For a lower-is-better performance metric $\gamma$, the oracle gain is the reduction obtained after revealing the candidate observation,
\begin{equation}
 a_{\mathrm{oracle}}(x;\mathcal{C}_k)
 = \gamma\!\left(p_\theta(\mathbf{y}_t\mid\mathbf{X}_t,\mathcal{C}_k),\mathbf{y}_t^{\star}\right)
 - \gamma\!\left(p_\theta(\mathbf{y}_t\mid\mathbf{X}_t,\mathcal{C}_k\cup\{(x,y_t^{\star}(x))\}),\mathbf{y}_t^{\star}\right).
\label{eq:oracle_gain}
\end{equation}
The first term is constant across candidates within a placement step, so maximising oracle gain is equivalent to minimising the post-augmentation metric.

For Joint NLL, the post-augmentation criterion is
\begin{equation}
 a_{\mathrm{JNLL}}(x;\mathcal{C}_k) \;=\; -\log p_\theta\!\left(\mathbf{y}_t^{\star}\mid\mathbf{X}_t,\mathcal{C}_k\cup\{(x,y_t^{\star}(x))\}\right).
\label{eq:joint_nll}
\end{equation}

For Gaussian marginals, the pointwise negative log likelihood is
\begin{equation}
 \ell_i(x,t) \;=\; \frac{1}{2}\log\!\left(2\pi\sigma_{t,i}^2(x)\right)
 + \frac{\left(y_{t,i}^{\star}-\mu_{t,i}(x)\right)^2}{2\sigma_{t,i}^2(x)}.
\end{equation}
Marginal NLL is the spatial mean of the pointwise NLL,
\begin{equation}
 a_{\mathrm{MNLL}}(x;\mathcal{C}_k) \;=\; \frac{1}{|\mathbf{X}_t|}\sum_{i\in\mathbf{X}_t}\ell_i(x,t).
\label{eq:marginal_nll}
\end{equation}

RMSE compares the predictive mean to the CAMS reference field,
\begin{equation}
 a_{\mathrm{RMSE}}(x;\mathcal{C}_k) \;=\;
 \sqrt{\frac{1}{|\mathbf{X}_t|}\sum_{i\in\mathbf{X}_t}\left(y_{t,i}^{\star}-\mu_{t,i}(x)\right)^2}.
\label{eq:rmse}
\end{equation}

Scores are computed per task time and then averaged across evaluation times. Joint MI, Marginal MI, and Mean Var are maximised. Joint NLL, Marginal NLL, and RMSE are minimised and are therefore negated when inserted into the maximisation form of Eq.~\ref{eq:greedy_generic}.

\subsection{Full acquisition-function placement comparison}
\label{app:placement_comparison}
\begin{figure}[H]
  \centering
  \includegraphics[width=\linewidth]{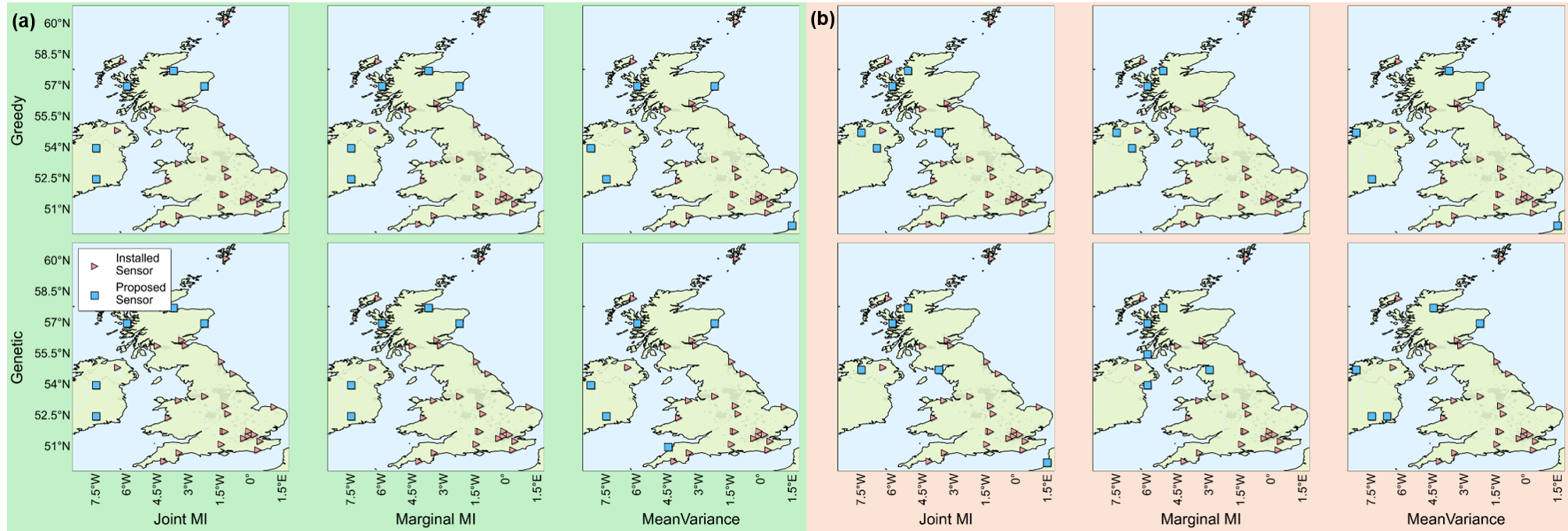}
    \caption{Final five-sensor placement maps for the acquisition criteria. (a) three model-based acquisition criteria and, (b) their corresponding oracle counterparts. Columns identify the acquisition criterion and rows compare greedy and genetic search.}
  \label{fig:app_placements}
\end{figure}

Figure~\ref{fig:app_placements} compares the final five-sensor networks produced by greedy and genetic search. The left panel shows the non-oracle acquisition criteria (Joint MI, Marginal MI, and Mean Var), while the right panel shows their corresponding oracle variants. The two solvers produce similar broad geographic coverage for several criteria but do not necessarily select identical grid cells. This is consistent with the sequential trajectories in Figure~\ref{fig:main_summary}(a), where different search paths retain comparable reference-based performance even when the final sensor coordinates differ.

\subsection{Acquisition-function agreement across time}
\label{app:weather_dependence}
\begin{figure}[H]
  \centering
  \includegraphics[width=\linewidth]{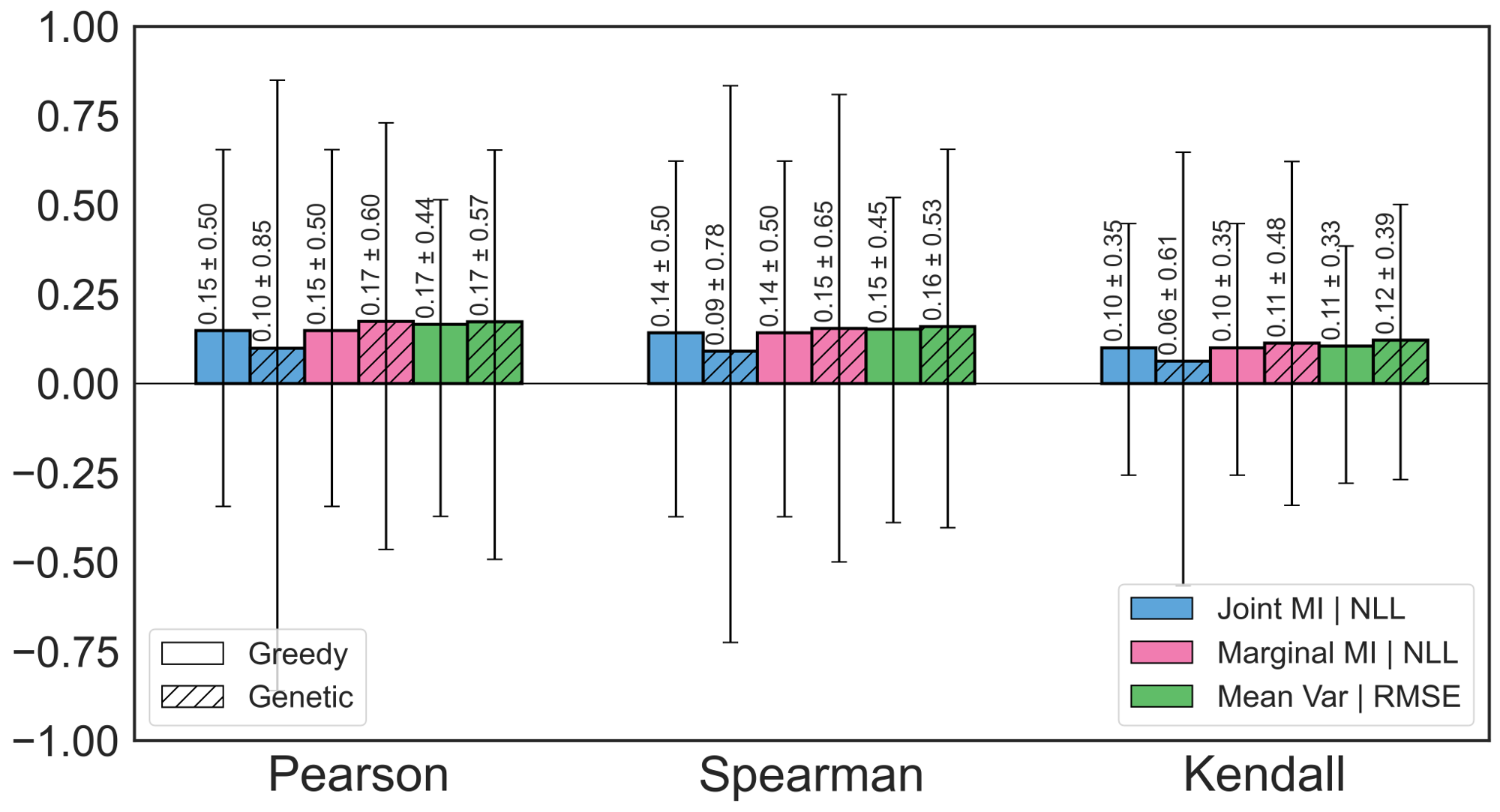}
  \caption{Correlation between paired model-based and oracle acquisition functions. The three pairs are Joint MI with Joint NLL, Marginal MI with Marginal NLL, and Mean Var with RMSE.}
  \label{fig:app_corr_summary}
\end{figure}

Figure~\ref{fig:app_corr_summary} summarises all-date agreement between the three model-based and oracle pairs for Pearson, Spearman, and Kendall coefficients. The mean correlations remain modest and temporal dispersion is large. Pearson means range from 0.10 to 0.17 across pair and solver combinations. Spearman means range from 0.09 to 0.16, while Kendall means range from 0.06 to 0.12. The standard deviations are larger than the means in many cases, indicating that acquisition-landscape alignment varies strongly between evaluation dates rather than remaining uniformly high through time.

\begin{figure}[H]
  \centering
  \includegraphics[width=\linewidth]{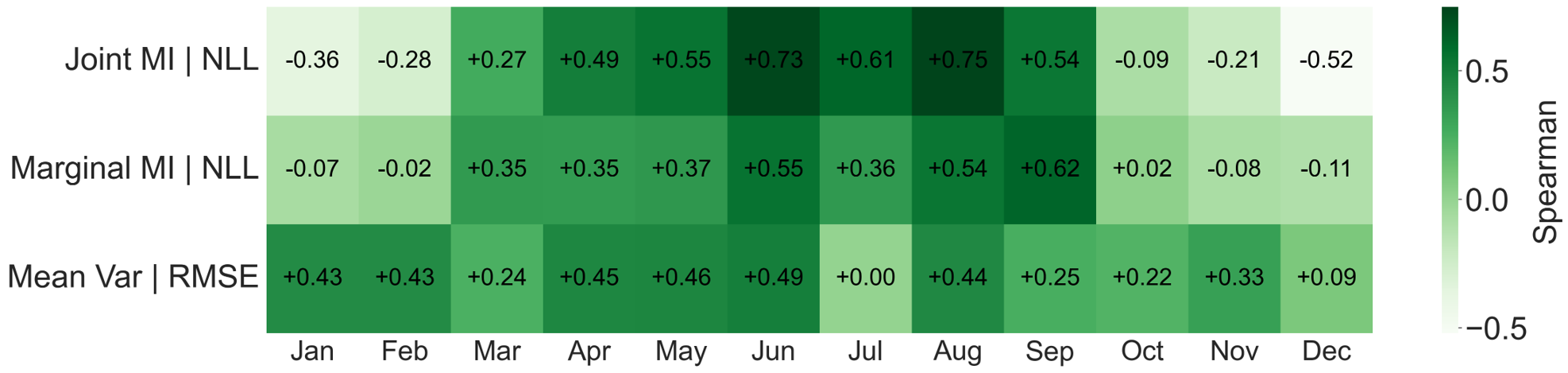}
  \caption{Monthly mean Spearman correlation between paired model-based and oracle acquisition functions.}
  \label{fig:app_monthly}
\end{figure}

Figure~\ref{fig:app_monthly} resolves the Spearman correlations by month. Joint MI with Joint NLL is negative in January ($-0.36$), February ($-0.28$), October ($-0.09$), November ($-0.21$), and December ($-0.52$), but is positive from March through September and peaks at 0.75 in August. Marginal MI with Marginal NLL is slightly negative in January ($-0.07$) and February ($-0.02$), becomes positive from March through October, peaks at 0.62 in September, and returns to $-0.08$ and $-0.11$ in November and December. Mean Var with RMSE remains non-negative throughout the year, ranging from 0.00 in July to 0.49 in June and reaching 0.44 in August. These contrasting seasonal patterns are consistent with the wind-regime dependence reported in Figure~\ref{fig:weather_main} and reinforce that different uncertainty criteria respond differently to changing transport conditions.

\subsection{Genetic optimiser details}
\label{app:ga_details}
The genetic solver represents each proposed sensor location by a two-gene chromosome $[i_{\mathrm{lat}},i_{\mathrm{lon}}]$, where the genes are integer indices on the latitude and longitude axes of the CAMS grid. The search space is the complete rectangular grid rather than a pre-filtered list of land candidates. For the UK experiment, the grid contains 15 latitude indices and 14 longitude indices, giving 210 possible coordinate pairs, of which 70 satisfy the land-sea mask criterion $\mathrm{lsm}>0.5$. A chromosome is feasible when both indices lie within the grid bounds, the corresponding cell satisfies the land mask, and the location has not already been selected in an earlier sequential placement step. Masked or previously selected locations are assigned an objective penalty with magnitude $10^{6}$, with the sign chosen according to whether the acquisition criterion is minimised or maximised, non-finite acquisition values receive the same unfavourable penalty. Feasible coordinate pairs are mapped to the compact land-candidate array used by DeepSensor for acquisition evaluation.

The genetic algorithm uses a population of 15 chromosomes and runs for 50 generations at each placement step. Five parents are selected for mating and the best three chromosomes are retained through elitism, corresponding to 20\% of the population. Uniform crossover is applied with probability 0.5, allowing latitude and longitude genes to be inherited independently from different parents and thereby creating new spatial combinations. Random mutation is applied with probability 0.2 per gene and changes coordinate indices within their allowed ranges. Since, crossover and mutation operate on the two spatial genes, they can generate both feasible land locations and infeasible combinations such as sea cells. Therefore, infeasible combinations are rejected through the fitness penalty rather than removed from the genetic search space.

To reduce computational cost, evaluations of $a(x;\mathcal{C}_k)$ are cached within each placement step. For acquisition functions that support parallel evaluation, the complete feasible score surface is computed once and the genetic algorithm subsequently queries this table. For other acquisition functions, evaluation is lazy: only feasible coordinate pairs actually visited by the genetic search are evaluated, while repeated visits return the cached score without additional model inference. This combination of structured two-gene search, feasibility penalisation, lazy evaluation, and caching produces the runtime reductions shown in Figure~\ref{fig:main_summary}(c).

\subsection{Specifications of the compute system}
\label{app:specific}
All computations for this study were performed on a workstation equipped with an Intel\textsuperscript{\textregistered} Core\textsuperscript{\texttrademark} i7-8850H processor (6 cores, 12 threads, 2.6 GHz base frequency, up to 4.3 GHz turbo), 32 GB of DDR4 memory (2 $\times$ 16 GB, 2667 MHz), and a 512 GB NVMe Gen 4 solid-state drive. GPU acceleration was handled by an NVIDIA Quadro P1000 GPU with 4 GB of GDDR5 VRAM. The software environment was executed on Windows 10 Pro (64-bit) with DeepSensor v0.1.5 and pyGAD v3.7.0~\cite{gad2023pygad} for model implementation and computational experiments.

\end{appendices}

\end{document}